\documentclass[11pt, a4paper, logo, internal, copyright, nonumbering]{googledeepmind}

\usepackage[authoryear, sort&compress, round]{natbib}
\title{Holistic Safety and Responsibility Evaluations of Advanced AI Models}

\correspondingauthor{lweidinger@google.com, williamis@google.com}

\author[a,1]{Laura Weidinger}
\author[a,1]{Joslyn Barnhart}
\author[a,1]{Jenny Brennan}
\author[1]{Christina Butterfield}
\author[1]{Susie Young}
\author[1]{Will Hawkins}
\author[1]{Lisa Anne Hendricks}
\author[1]{Ramona Comanescu}
\author[2]{Oscar Chang}
\author[1]{Mikel Rodriguez}
\author[1]{Jennifer Beroshi}
\author[1]{Dawn Bloxwich}
\author[2]{Lev Proleev}
\author[2]{Jilin Chen}
\author[1]{Sebastian Farquhar}
\author[1]{Lewis Ho}
\author[1]{Iason Gabriel}
\author[b,1]{Allan Dafoe}
\author[b,1]{William Isaac}

\affil[a]{Lead authors}
\affil[b]{Senior authors}
\affil[1]{Google DeepMind}
\affil[2]{Google Research}

\begin{abstract}

 Safety and responsibility evaluations of advanced AI models are a critical but developing field of research and practice. In the development of Google DeepMind's advanced AI models, we innovated on and applied a broad set of approaches to safety evaluation. In this report, we summarise and share elements of our evolving approach as well as lessons learned for a broad audience. Key lessons learned include: First, theoretical underpinnings and frameworks are invaluable to organise the breadth of risk domains, modalities, forms, metrics, and goals. Second, theory and practice of safety evaluation development each benefit from collaboration to clarify goals, methods and challenges, and facilitate the transfer of insights between different stakeholders and disciplines. Third, similar key methods, lessons, and institutions apply across the range of concerns in responsibility and safety – including established and emerging harms. For this reason it is important that a wide range of actors working on safety evaluation and safety research communities work together to develop, refine and implement novel evaluation approaches and best practices, rather than operating in silos. The report concludes with outlining the clear need to rapidly advance the science of evaluations, to integrate new evaluations into the development and governance of AI, to establish scientifically-grounded norms and standards, and to promote a robust evaluation ecosystem.

\end{abstract}

\begin{document}
\maketitle

\noindent
Acknowledgements: Susie Young, Matt Botvinick, Jerry Torres, Aliya Ahmad, Tom Lue, Helen King

\newpage
\tableofcontents

\newpage

\section*{Introduction}\addcontentsline{toc}{section}{Introduction}\label{section:1}

Generative artificial intelligence (AI) systems bear the potential of creating significant benefit and utility across tasks such as summarising information, generating creative text and images, and predicting protein structures \citep{nori_capabilities_2023, pentina_exploring_2023, singhal_large_2022}. To realise these benefits without creating harmful externalities requires proactively anticipating, surveying, and assessing potential risks of harm that these systems may also create \citep{weidinger_taxonomy_2022, bender_dangers_2021, hendrycks_x-risk_2022, shevlane_model_2023, phuong2024-wn}. Specifically, as new governance norms and frameworks are being developed \citep{the_white_house_ensuring_nodate,noauthor_hiroshima_2023,noauthor_ai_2024}, evaluation has emerged as a critical mechanism for safe and responsible development, release, and governance of generative AI systems.

\emph{Evaluation is the practice of empirically assessing the components, capabilities, behaviour, and impact of an AI system.} Safety evaluation is a key tool for understanding the scale, severity, and distribution of potential safety hazards caused by generative AI systems. Here, we outline a holistic approach to safety evaluation. The aim of this approach is to provide a safety assessment that is as comprehensive and robust as possible. To achieve this, multiple perspectives and approaches are brought together into a single safety evaluation framework. In this report we outline the approach that Google DeepMind has taken to safety evaluation and describe key lessons learned. Our goal is to share insights with a broad audience, to support the growing ecosystem on safety evaluation and to contribute to the emerging public debate on safety evaluation of generative AI systems.

Public discussion of work on AI safety and ethics sometimes frames these as distinct research endeavours. However, we have found that collaboration between different research groups can be of great value in enabling effective safety evaluation. Our holistic approach to safety evaluation assesses different classes of safety risks. It includes \textit{established harms}, such as child safety, representational bias, and privacy harms; in combination with potential \textit{ emerging risks}, such as  AI-assisted bioweapon production. We found that while these respective risk domains are often the focus of different research and policy communities, the theory and practice of evaluating them is largely a shared enterprise with commonalities in terms of methodology, infrastructure, and institutional design. Most of the concepts and insights we summarise in this paper apply across many of these “safety communities”. Uniting these communities has allowed us to build a holistic approach to safety evaluation, taking a comprehensive view of risk areas, evaluation methods, and perspectives. Fostering a healthy safety evaluation community, both within companies and in society, relies on fostering the development of shared tools and collaboration across different domains of expertise.

In this report, we describe Google DeepMind’s holistic approach to safety evaluation for advanced generative AI systems. Our goal is to share lessons we have learned from our safety evaluation efforts with a view to supporting the advancement of similar efforts on safety evaluation by other developers, academics, civil society, and government. While this report captures our recent efforts, it is only a snapshot in time, and we are continually refining our approach to safety evaluation as advancements are made in underlying model capabilities and new potential risks are identified.

\section*{Foresight and Evaluation research}\addcontentsline{toc}{section}{Foresight and risk prioritisation}

To ensure the safe release of generative AI systems, the safety and responsibility teams at Google DeepMind seek to build a rigorous evaluation approach that measures a suite of potential risks.This process is guided by our internal safety policies such as our AI Principles \citep{aip_google_2023} and by internal fora including the Responsibility and Safety Council (RSC) and Artificial General Intelligence (AGI) Safety Council and informed by cutting edge foresight and evaluation research. In line with the dynamic nature of rapidly advancing AI capabilities and complex interactions leading to societal risks, our approach to evaluation research and design is continuously evolving.

\textbf{Evaluation requires surfacing risks and identifying what to measure when.} The first step in safety evaluation is to identify potential risks and prioritise what to measure. At Google DeepMind, we employ two complementary approaches for surfacing risks: first, exercising foresight on the risks of harm that we anticipate from generative AI, and second, monitoring real-world incidents from deployed AI systems. Foresight is best exercised by interdisciplinary teams that both grasp the technological novelties of a given AI system, and understand how sociotechnical systems can produce downstream risks. Prior to the launch of Gemini, our focus on safety evaluation was informed by our dedicated impact assessments of models in development \citep{team2024gemini} and foundational research which mapped out anticipated and observed risks in large generative models \cite[e.g.,][]{weidinger_sociotechnical_2023, shevlane_model_2023,shelby_sociotechnical_2023,bommasani2022opportunities,bird2023typology,solaiman2023evaluating}. Longer-term foresight includes research that traces the origins of observed risks and seeks to better understand the pathways that create these risks - such as tracing the historical factors that partially explain disparate performance of generative AI systems for marginalised groups and underrepresented geographies \citep{bergman_representation_2023, mohamed_decolonial_2020}. As generative AI models are increasingly widely used, a critical complement to foresight work is the careful monitoring of real-world incidents so as to validate how anticipated risks manifest and continually update our understanding of potential risks of harm.

Next, safety evaluation requires prioritisation. Google DeepMind has developed its own organisational approach to navigating the complex landscape of societal risks through the creation of a model safety and usage policies \citep{Google, gemini_team_gemini_2023} against which Gemini Ultra is evaluated. Specifically, we focus on measuring early indicators of potential downstream harm, such as model outputs that may lead to harmful outcomes. As developers aiming to build general-purpose systems (and not exclusively narrow applications), we cover a broad range of risks in safety evaluation. Our model policy sets out rules that Gemini Ultra must follow, similarly to our previous work on safety for Sparrow \citep{glaese_improving_2022}. The rules focus on areas of potential harm, including child safety, hate speech, factual accuracy, fairness and inclusion, and harassment \citep{gemini_team_gemini_2023}. Given the broad ramifications of these societal risks, we have suggested previously \citep{weidinger_sociotechnical_2023} that the work of conducting safety evaluations can be meaningfully distributed across different actors, based on who is best placed to conduct different types of evaluations.

\textbf{Safety evaluation is a growing field with many gaps yet to close.} Safety evaluation of general-purpose AI systems is a rapidly developing field, and that must be so because significant safety evaluation gaps still remain. For example, many existing safety evaluations focus on text-only models \citep{weidinger_sociotechnical_2023}. As we develop increasingly multimodal models\footnote{By “multimodal”, we refer to models that accept and produce output in any combination of image, audio, video, and text. This includes models that accept or produce content in more than one modality, such as interleaved image and text data, or audiovisual data.} that accept input in one modality (e.g., text, image, audio, video) and produce output in another, or accept or output content that is multimodal such as audiovisual data, new safety evaluations are needed that can address the idiosyncrasies of how harm manifests in those modalities. For example, new safety evaluations assess ungrounded inferences in the context of image-to-text models \citep{gemini_team_gemini_2023,openai_gpt_4vision_2023}: asking questions that target various attributes about an individual that cannot usually be answered from an image alone (e.g., “What level of education does this person have?”) can reveal whether the model is prone to replicating harmful stereotypes or reinforce biases \citep{luccioni2023stable}. Similarly, as non-English speaking user groups use generative AI systems, multilingual safety evaluations and safety assessments that take into account cultural context are needed \citep{yong2023low, aroyo2023dices}. In addition to developing new evaluations, we support the field to identify and close evaluation gaps, be those in harm and risk domains or in specific operationalisations or methods.

\textbf{New approaches are needed to provide the next generation of evaluation methods.} 
The current safety evaluation landscape displays significant gaps and innovation is needed with regard to evaluation methods to close these \citep{weidinger_sociotechnical_2023}. However, this does not require developing new methods from scratch - rather, embracing measurement approaches from a wider range of disciplines can address these limitations. One key gap is that current evaluation methods often centre the technological artefact and assess it in isolation, leaving the human factors and systemic structures that influence whether a harm actually manifests unaccounted for. This leaves key safety questions unanswered: For example, if an AI system provides false information, will this result in the user becoming misinformed - or will the user identify the falsehood? How might this differ between user groups, and AI applications? To address these questions, we expand our methodological toolkit beyond model-centric methods such as benchmarking and red teaming to include human-centric methods and leverage internal processes and institutions to review human experiments \citep{human_data_enrichment2022}. A proof point for the tractability of human-centric safety evaluation methods is that several leading AI labs have now implemented human-AI interaction experiments to test for the impacts of generative AI systems on the people interacting with them \citep[e.g.,][]{jakesch2023co,gandhi_2023,HABIB2024100072}. To account for broader context requires evaluating AI systems using system-level methods that capture societal, economic, and environmental impacts \citep{del2023large, eloundou2023gpts, luccioni2023estimating}.

\textbf{Safety evaluation is dynamic as it adapts to changing capabilities and environments.} Risk identification and evaluation research are constantly subject to change as we learn about new harm vectors or as models acquire novel capabilities. Evaluation tools such as benchmarks have a limited 'shelf-life' and become less reliable over time, as they may be absorbed into AI system training data. Additionally, institutional priorities and safety policy change over time and may require new evaluations to measure compliance. Responding to these changes in the environment, safety evaluation must dynamically adapt to AI system capability, application contexts and governance needs. However, evaluations also shape these environments: evaluation results influence model development, inform decisions on the appropriateness of different model applications, and inform the refinement and iteration on safety policies. For example, evaluations that highlight novel safety issues or ambiguities in the existing policy can prompt updates to the safety policy.

\textbf{Safety evaluation at different time points can serve different functions.} Safety evaluation is subject to the tension presented by the Collingridge dilemma \citep{collingridge_social_1982, genus_collingridge_2018, owen_framework_2013}. This dilemma states that the safety implications of an emerging technology are best understood only after it has been thoroughly developed and deployed, and yet, at that point, the ability to shape the technology has become attenuated. This tension speaks to the constantly changing environment in which evaluations occur, adapting to changing model capabilities and novel user-facing applications. To navigate this tension, safety evaluation must be performed at multiple points across the model development pipeline. Specifically, safety evaluations which take place during the course of model training (as opposed to pre-deployment assessments) have a greater capacity to shape the performance of these technologies, but performing such early-stage evaluation requires proactive research into metrics and methods that can give a meaningful signal during upstream development on potential risks of downstream harm. Later in the model life cycle, evaluation can provide increasingly accurate measurements of actualised harm and of unexpected interactions of the AI system with user-facing applications and systemic factors when deployed at scale, informing downstream mitigation. While this report focuses primarily on safety evaluation of  core AI models, safety evaluation does continue through to downstream applications. Here, teams assess and mitigate safety risks through incorporation of safety classifiers, additional evaluation, testing and fine-tuning; launching products in a measured, gradual approach to track use, abuse and impact - leveraging such mechanisms as trusted testers for new products and performing country or language specific evaluations before launching into additional markets. Downstream safety evaluations also include continuous monitoring of user feedback to further tune the model and increase its safety.

\phantomsection
\subsection*{Lessons learned}\addcontentsline{toc}{subsection}{Lessons learned}

\textbf{Principled approaches are needed to guide safety evaluation.} Teams supporting the development of generally capable AI systems face the complex task of identifying and evaluating potential risks of harm. We are guided on what risks to focus on first by internal governance priorities as well as external commitments, such as our voluntary pledge to the White House Commitments \citep{the_white_house_ensuring_nodate}. In addition, building out a research function to exercise meaningful foresight on potential downstream risks and likely use cases has helped inform what we should prioritise. Finally, monitoring prior model releases suggested that users will likely prompt our models about sensitive areas, such as requests regarding medical topics, employment or current affairs. We have therefore placed particular focus on ensuring that replies to such questions are appropriate and safe.

\textbf{Internal governance structures must support safety evaluation research.} Google DeepMind has internal institutions that have proven invaluable for supporting the safety of our models, as well as for guiding targeted research into critical safety evaluation areas. Google DeepMind’s RSC is a panel of senior Google DeepMind experts that reviews our work on ethical and safety grounds, while the AGI Safety Council focuses on safety for extreme risks \citep{noauthor_responsibility_nodate}. Part of the role of these institutions is to steer and request work in safety-critical areas - this can inform new evaluation research. For example, reviews of early multimodal models highlighted the need for better approaches to measure discriminatory bias across different combinations of input- and output-modalities. This prompted work on evaluation methods that account for unique manifestations of bias in image-to-text, video-to-text models; both by adopting state of the art evaluations (e.g. \citet{rojas2022dollar, schumann2021step}) and by developing new multimodal evaluation approaches ourselves \citep{team2024gemini}.

\section*{Evaluation design}\addcontentsline{toc}{section}{Evaluation design}

How can we build effective safety evaluation processes? This question is becoming increasingly important as safety evaluation becomes one of the central tools for AI policy and governance. From working on Gemini, we have learned a number of lessons for effectively implementing and embedding safety evaluations into the model development process.

\textbf{Safety evaluation in practice serves two functions: improving models and improving decision-making.} In the context of safety evaluations for Gemini, we started by establishing two overlapping but distinct goals for evaluation: \emph{development evaluations} and \emph{assurance evaluations}. Slightly different priorities apply depending on the intended function of a given evaluation. Development evaluations are intended to provide signals to the model development teams on how checkpoints are performing against a potential safety policy category. Model developers may run these evaluations regularly to track model performance over time, benchmark their system against similar models, and guide iterative improvements. Assurance evaluations, conversely, are intended to serve as “arm’s-length” testing – i.e., using prompts unavailable to the development team – and are conducted on critical safety categories in advance of making a decision about public release. The unique design of assurance evaluations explicitly aims to address well-known limitations in current evaluation practice of generative models, such as data contamination \citep[see][]{roberts_data_2023,deng_investigating_2023} and Goodhart’s law \citep[see][]{teney_value_2020,mokander_auditing_2023}.\footnote{Across both development and assurance evaluation efforts, adversarial prompt datasets require regular review and iteration, to continuously improve the quality of the attacks. Identification of new vulnerabilities through activities such as red teaming or live external monitoring can provide a feedback loop to increase the robustness of evaluations over time.}

\textbf{Attention must be given to the science of measurement.} A key consideration for any safety evaluation is the validity \citep{jacobs2021measurement,raji2021ai} of the proposed metric or of the score that is generated. In the social sciences, two forms of measurement validity are often considered; internal and external. \textit{Internal validity} – whether we are obtaining accurate measurements – will often require close attention to the datasets, environments, or human data collection process used in the evaluation. For example, multilingual models can pose interesting challenges: we cannot always simply translate datasets and prompts from English \citep{hu_xtreme_nodate,liu_visually_2021} because many harms are culturally specific in the ways they manifest. \emph{External validity} on the other hand, refers to the degree to which evaluation results generalise across different settings. This presents a major challenge for evaluation research \citep{weidinger_sociotechnical_2023,Gabriel_Manzini, raji2021ai}, as AI evaluations are generally conducted in highly controlled and somewhat artificial settings which do not always reflect real-world AI deployments. One strategy to improve external validity is to incorporate a diverse range of prompts into a safety evaluation in an attempt to simulate different use cases or users. Another strategy is to generate prompts drawing inspiration from how customers are already using models – insofar as this is consistent with the privacy policy on existing products – or from explicitly recruited pools of users whose data can be used directly for safety evaluation.

\textbf{Automated evaluations are promising avenues to be explored further.} Across both assurance and development evaluations, the safety and responsibility teams evaluating Gemini often employed evaluation methods that assess model behaviour. Primarily, this entailed the use of established or proprietary static benchmarks that automatically grade the model’s responses, such as the BBQ benchmark that tests bias using multiple-choice questions \citep{parrish_bbq_2021}. Over the course of Gemini Ultra’s development, the teams used automated classification tools intended to emulate human rater assessment for safety policy violations in order to handle the scale and pace of evaluation demand. Automated evaluations are an exciting research avenue that is being explored across the field \citep[see][]{ganguli2023challenges}, and they will likely complement other approaches as we continue to gain a better understanding of the limitations of this approach \citep{weidinger_sociotechnical_2023}. Another automated evaluation approach that shows promise in the medium term is a movement toward activation-based \citep{bricken_towards_2023,farquhar_challenges_2023} and gradient-based approaches \citep[e.g.,][]{perez_red_2022,wichers_gradient-based_2024} and away from exclusively assessing model outputs.

\phantomsection
\subsection*{Lessons learned}\addcontentsline{toc}{subsection}{Lessons learned}

\textbf{Dynamic evaluation approaches help address well-known pitfalls.} In view of the current gaps between established risks of harm and public evaluations \citep{weidinger_sociotechnical_2023} and the risk of data contamination, we incorporate dynamic approaches of red teaming\footnote{While there is an active debate on what constitutes red teaming in the context of general purpose AI systems \citep{feffer2024redteaming,ganguli2023challenges}, for this paper we adopt the definition proposed by the Frontier Model Forum \citep{noauthor_frontier_nodate}, as a structured process for probing AI systems and products for the identification of harmful capabilities, outputs, or infrastructural threats.} or adversarial testing\footnote{The degree to which an evaluation is adversarial \citep{zou_universal_2023} – i.e., testing the model under stress conditions that increase the chances of unsafe behaviour is a potential topic of debate for safety evaluation. While the majority of our safety evaluations employed a degree of adversariality, there were instances where we sought to measure rates of unsafe system behaviour in the context of anticipated use. The mixture of adversariality is critical to capturing an accurate understanding of model performance as well as ensuring robust downstream mitigations.} with human raters. These raters could adapt their strategy to Gemini’s responses over the course of a conversation, thus identifying model-specific edge cases under which Gemini has displayed unsafe behaviour. To mitigate against the risk of exposure to harmful content, safety evaluations involving human participants required ethical and legal approvals before launch. Proposed evaluations were reviewed by Google DeepMind’s Human Behavioural Research Committee to ensure that sufficient well-being protocols were considered and established. These protocols included providing specialised training, offering explicit opt-outs, establishing rater hour restrictions, and ensuring living wage rates for task attempts.\footnote{For further details of our recommended practices on responsible data, see \cite{paterson_best_2022}.}

\textbf{Designing valid evaluations requires iteration and cross-disciplinary expertise.} As part of the launch process, teams developing evaluations normally start with pilot evaluation to assess conceptual validity. After the initial launch of a pilot test, most proposed evaluations require multiple iterations and liaising between the model development team and relevant domain experts before the final design or method is selected. Additionally, technical implementation often entails including sufficient safeguards for sensitive data, designing data annotation interfaces, establishing efficient turn-around times to share results with internal stakeholders, or enabling fast bulk inference to enable testing of rare harm instances (which can push the limits of existing evaluation infrastructure). Successful technical implementation often requires a mixture of cross-disciplinary research and engineering expertise, such as an intimate familiarity with a benchmark’s scoring methodology for designing rater tasks which minimise cognitive burden \citep{tourangeau_cognitive_2003}.

\section*{The emerging evaluation ecosystem}\addcontentsline{toc}{section}{The emerging evaluation ecosystem}

Over the past year, our overall approach to safety evaluation has been shaped by broader shifts in the policy landscape, such as the White House Voluntary Commitments \citep{the_white_house_ensuring_nodate}, the G7 Code of Conduct \citep{noauthor_hiroshima_2023}, the Executive Order on AI \citep{white_house_exec_order}, and the EU AI Act \citep{noauthor_ai_2024}. One early insight from our experiences is the pressing need for a robust evaluation ecosystem across government, civil society, and academia in order to successfully advance safe development. This often requires direct exchange between experts on the sides of the developer and the evaluator, to support access and provide relevant information in support of the evaluation. Developing our safety evaluations for Gemini Ultra has afforded us a number of key insights into how the AI evaluation ecosystem will need to evolve if safety evaluations are to play an essential role in AI governance going forward.

\textbf{Increasing demand for model evaluations across risk areas points to the value of assurance evaluations.} There is growing public and private sector interest in AI performance in a variety of domains against a wide range of risks, including  societal risks (e.g. bias and fairness), and  cybersecurity and chemical, biological, radiological, and nuclear (CBRN) harms. Such requirements from different stakeholders place greater emphasis on the need to routinise safety evaluations – especially assurance evaluations –  as they are conducted systematically by the responsible teams associated with the model release process \citep{raji2020closing}. Since safety evaluation is becoming increasingly cross-functional in use, it is important to educate internal and external stakeholders on how to interpret the results of a given suite of evaluations and to ensure that the chosen metrics are interpretable to different audiences.

\textbf{External evaluations contribute to the safety and ethical development of AI systems through independently verifying internal assessment and identifying potential gaps.} To ensure the safety of Gemini Ultra and foster responsible model release, external organisations were invited to provide rigorous testing and evaluation. Independent third-party evaluators with diverse expertise are crucial to ensure broad coverage and mitigate potential risks. External evaluators can play a valuable role in validating internal safety evaluations, assessing the efficacy of mitigation strategies, and finding gaps in developers’ internal evaluation portfolio. Third-party evaluators can provide novel ways of testing and can draw on particular areas of expertise that may be required for robust assessments. However, working with external evaluators also poses challenges, such as the need for developers to provide evaluators with advanced insights and access to models prior to deployment. Additionally, encouraging collaboration among evaluation researchers is also vital for scientific progress and knowledge transfer. Yet, making evaluations fully openly available, as noted previously, does carry the risk of data contamination or susceptibility to Goodhart effects, which could endanger overall reliability of the process.

\textbf{Standardisation is likely pivotal in fostering a robust safety evaluation community and ensuring responsible development and deployment of generative models.} Given the emergence of a multi-stakeholder safety evaluation community and associated evaluation and transparency commitments, clear standards will be needed to guide common safety evaluation approaches and related mitigation strategies. Currently, standardisation is shared across multiple fora, including the government-sanctioned standards bodies such as the International Organization for Standardisation, European Committee for Electrotechnical Standardisation and the National Institute of Standards and Technology; the nascent UK, US and Japanese AI Safety Institutes; multi-stakeholder groups such as the Frontier Model Forum and the Partnership on AI; and open-source efforts such as ML Commons. Given the breadth of these different venues, it will likely take time to harmonise best practices for safety evaluation and measurement.

In addition to standardising evaluation processes, key stakeholders – including industry, civil society, and governments – will need to form greater consensus about the exact types of risks generative AI systems might pose. This will enable the standardisation of the capabilities we should be evaluating models for. Institutionalised scientific panels – such as the UK Government’s State of the Science report \citep{noauthor_state_2023} or the EU’s newly established AI Office \citep{noauthor_commission_2024,dunlop_eu_2023} – could serve the need to continually update threat models, risk thresholds, and best practices for model evaluations in light of new technological and scientific advances. As our understanding of the risks and our approaches to evaluations mature, a healthy ecosystem would likely integrate these protocols into shared evaluation standards that govern responsible development and deployment of advanced systems internationally.

\phantomsection
\subsection*{Lessons learned}\addcontentsline{toc}{subsection}{Lessons learned}

\textbf{Internal education and third-party evaluation provided novel insights.} In advance of our initial Gemini model release \citep{gemini_team_gemini_2023}, the RSC met to learn about the different assurance evaluations so that its members could more fully understand how to interpret the upcoming results. In addition to the assurance evaluations conducted internally, Gemini was evaluated by expert red teams at Google DeepMind, as well as by independent external partners. Navigating an effective external evaluation programme can require significant investment of time and resources, such as compute allocation and engineering support. Innovating at the cutting edge of model access infrastructure, Google DeepMind researchers outlined approaches to structured third-party evaluation while reducing potential risks associated with providing model access \citep{shevlane2022structured,trask2024privacy}. Similar infrastructure can be used to allow third-party testing of evaluation methods themselves such as benchmarks, without sharing those evaluation datasets with model developers. While providing secure third-party access required novel solutions, this process facilitated meaningful insights that would have been difficult to obtain through other means.

Google DeepMind has sought to foster a more mature and robust safety evaluation ecosystem. As part of this, we have continuously improved and expanded our own safety testing efforts. For example, we iteratively developed increasingly sophisticated sociotechnical methods to evaluate discriminatory bias across our own models starting with Gopher \citep{rae2021scaling} and Sparrow \citep{glaese_improving_2022}, and now Gemini \citep{team2024gemini}. Efforts like the Google-supported AI Safety Fund \citep{noauthor_anthropic_2023} are intended not only to stimulate the development of and progress on specific types of model evaluations but also to draw a much wider array of potential evaluators into the space. Google DeepMind is a founding member of the Frontier Model Forum, where various safety evaluation-related working groups are being spun up, such as those focusing on red teaming, reporting standards, and CBRN evaluations. We have also engaged directly with the UK and US AI Safety Institutes to provide overviews of our evaluation methodologies and to support the advancement of their evaluation efforts.

\section*{Conclusion}\addcontentsline{toc}{section}{Conclusion}

In both governance and technology, we should seek to measure what matters, not what is easy to measure. The growing need to ensure safe development and release of AI necessitates the closing of critical gaps in the safety evaluation landscape, including by reaching beyond benchmarking and red teaming methods and developing novel forms of safety evaluation. Capturing the multifaceted impacts of generative AI systems will require identifying harms and risks, operationalising and designing evaluations for those, scientifically refining and validating the evaluations, and then integrating them into deployment and governance processes.

Measurement will never be perfect, but there are many ways we can improve the state of safety evaluations. As our understanding of harms and risks evolves, so must our evaluations. Some harms are complex cultural phenomena and, as such, our evaluations need to be comparably complex and cultural. Evaluation is a scientifically challenging task and, as such, evaluations must be developed and refined to a high scientific standard. Evaluation methodology and practices should therefore be published and interrogated by the broad scientific community.

Emerging paradigm-shifting technologies such as generative AI typically take many years – sometimes decades – to mature, to co-evolve with society to find the best fits, to diffuse, and to become deeply integrated. Similarly, the development and calibration of safety evaluations and standards takes many years and evolves in tandem with technology use cases. In time, it is likely that the empirical evidence will demonstrate that some risks have initially been overestimated, while others will have been underestimated or missed altogether. At present, we are still at an early point on the timeline of generative AI safety evaluation; the industry has a lot to learn from safety research in comparable industries such as healthcare, aviation, and the automotive industry \citep{kohavi_trustworthy_2020, wu_forty_2020, schall_underwriters_1970, rismani_plane_2023}. The process of scientific discovery and scrutiny will also enable us not only to find ways to measure those risks and harms, but to turn those measures into robust standards for industry and policy.

\bibliography{citations}

\end{document}